\documentclass{SCIS2021}

\usepackage{hyperref}       
\hypersetup{
    colorlinks=true,
    linkcolor=blue,
    filecolor=blue,
    urlcolor=blue,
    citecolor=cyan,
}
\usepackage{url}            
\usepackage{booktabs}       
\usepackage{amsfonts}       
\usepackage{nicefrac}       
\usepackage{microtype}      

\usepackage{graphicx}
\usepackage{url}
\usepackage{dblfloatfix}
\usepackage{caption}
\usepackage{epstopdf}

\begin{document}
\ArticleType{RESEARCH PAPER}
\Year{2020}
\Month{}
\Vol{}
\No{}
\DOI{}
\ArtNo{}
\ReceiveDate{}
\ReviseDate{}
\AcceptDate{}
\OnlineDate{}

\title{Onfocus Detection: Identifying Individual-Camera Eye Contact from Unconstrained Images}{Onfocus Detection: Identifying Individual-Camera Eye Contact from Unconstrained Images}

\author[1]{Dingwen Zhang$\dag$}{}
\author[2,4]{Bo Wang$\dag$}{}
\author[1]{Gerong Wang$\dag$}{}
\author[1]{Qiang Zhang*}{{qzhang@xidian.edu.cn}}
\author[1]{Jiajia Zhang}{}
\author[3]{\\Jungong Han*}{{jungonghan77@gmail.com}}
\author[2]{Zheng You}{}

\AuthorMark{D. Zhang}

\AuthorCitation{D. Zhang, B. Wang, Ge. Wang, et al}

\contributions{Equal contributions.}

\address[1]{School of Mechano-Electronic Engineering, Xidian University, Xi'an, 710071, China}
\address[2]{State Key Laboratory of Precision Measurement Technology and Instruments, Tsinghua University, Beijing, 100084, China}
\address[3]{Computer Science Department, Aberystwyth University, Ceredigion, SY23 3FL, UK.}
\address[4]{Beijing Jingzhen Medical Technology Ltd., Beijing, 100084, China}

\abstract{Onfocus detection aims at identifying whether the focus of the individual captured by a camera is on the camera or not. Based on the behavioral research, the focus of an individual during face-to-camera communication leads to a special type of eye contact, i.e., the individual-camera eye contact, which is a powerful signal in social communication and plays a crucial role in recognizing irregular individual status (e.g., lying or suffering mental disease) and special purposes (e.g., seeking help or attracting fans). Thus, developing effective onfocus detection algorithms is of significance for assisting the criminal investigation, disease discovery, and social behavior analysis. However, the review of the literature shows that very few efforts have been made toward the development of onfocus detector due to the lack of large-scale public available datasets as well as the challenging nature of this task. To this end, this paper engages in the onfocus detection research by addressing the above two issues. Firstly, we build a large-scale onfocus detection dataset, named as the OnFocus Detection In the Wild (OFDIW). It consists of 20,623 images in unconstrained capture conditions (thus called ``in the wild'') and contains individuals with diverse emotions, ages, facial characteristics, and rich interactions with surrounding objects and background scenes. On top of that, we propose a novel end-to-end deep model, i.e., the eye-context interaction inferring network (ECIIN), for onfocus detection, which explores eye-context interaction via dynamic capsule routing. Finally, comprehensive experiments are conducted on the proposed OFDIW dataset to benchmark the existing learning models and demonstrate the effectiveness of the proposed ECIIN. The project (containing both datasets and codes) is at \url{https://github.com/wintercho/focus}.}


\maketitle

\section{Introduction}
\begin{figure}[t]
    \centering
    \includegraphics[width=1\textwidth]{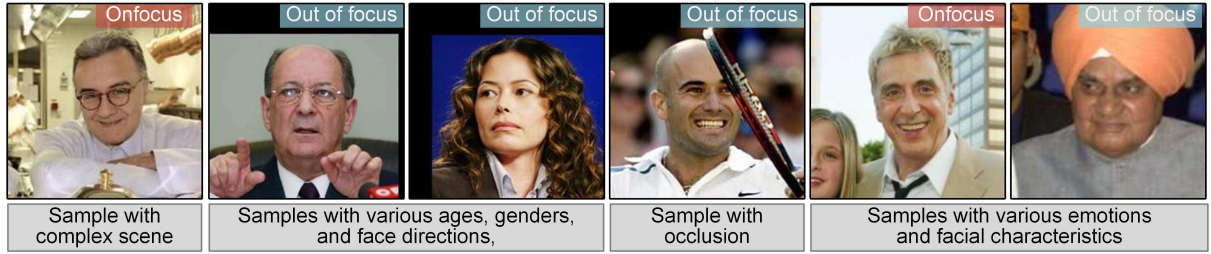}
    \caption{Examples to illustrate the onfocus detection task and the challenge scenarios.} \label{1}
\end{figure}

In this work, we define onfocus detection as the task to identify whether the focus of the individual (human or animal) captured by a camera is on the camera or not, which is performed on images captured in unconstrained imaging conditions (so-called ``in the wild'') as shown in Figure~\ref{1}. 
The focus of an individual during face-to-camera communication leads to individual-camera eye contact, which is a special type of eye contact and plays a crucial role in social communication. Such eye contact could reflect irregular individual status (e.g., lying~\cite{1} and suffering mental disease~\cite{2}) or special purposes (e.g., seeking help~\cite{3} and attracting fans~\cite{4}).  To this end, onfocus detection tends to be instrumental to a wide range of real-world applications in the fields of criminal investigation, disease discovery, social behavior analysis and it also has value for facilitating robots for human-machine interaction and communication.

In spite of its academic value and practical significance, onfocus detection also presents
great challenges in unconstrained capture conditions due to the complex image scenes, unavoidable occlusion, diverse face directions and emotions, various facial features (especially for the features of the eyes), and imagery factors like blur, over-exposure, etc. Some examples to illustrate the challenges of onfoces detection in the wild can also be referred to in Figure~\ref{1}.

According to the existing literature, few efforts have been made toward the onfocus detection problem, leaving most of the aforementioned challenges under-studied\footnote{It worth mentioning that the method in \cite{39} aims to estimate whether a certain image pixel (from the whole image scene) is at the camera’s focus distance or not whereas our study tries to estimate whether the person within the image is staring at the camera or not.}. Besides, there is not a publicly available dataset for this task. To advance this research field, this paper establishes a large-scale onfocus detection dataset called OnFocus Detection In the Wild (OFDIW). OFDIW consists of 20,623 unconstrained images which are collected from the LFW dataset~\cite{5} and the Oxford-IIIT Pet dataset~\cite{6}, enabling us to study the individual-camera eye contact for both humans and pets. It is worth mentioning that human-pet communication becomes prevalent in our daily life. Many people raise pets and some companies raise pets and use them to help people relieve stress. However, due to the barrier in language, human-pet communication heavily relies on eye contact. Thus, we think involving the onfocus study of pets might be beneficial and we thus choose two of the most common pets, i.e., dog and cat, for our study. Other pets will be studied in the future. Each image in this dataset is labeled with the corresponding ground-truth, i.e., either onfocus or out of focus, by human annotators. With the large-scale images and ground-truths, OFDIW can benefit methods based on the current deep learning models. Besides, the individuals contained in OFDIW have good diversity in emotions, ages, facial characteristics, and rich interactions with surrounding objects and background scenes. The samples also satisfy the imbalanced data distribution. All these factors make OFDIW closer to real-world scenarios. This dataset will be publicly accessible to other researchers to promote development in this area.

\begin{figure}[t]
	\centering
	\includegraphics[width=1\textwidth]{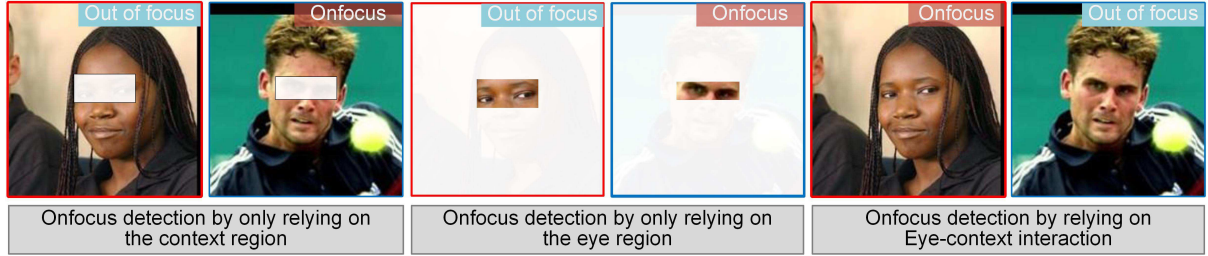}
	\caption{Illustrate the importance of exploring eye-context interaction for onfocus detection. From the examples, we can observe that onfocus detection by only relying on either the context region (the first block) or the face region (the second block) may lead to incorrect results while considering the eye-context interaction (the third block) would obtain the correct results. The images framed with the same color are the same sample. The sample with the red frame shows the interaction of the eye region and the face direction, while the sample with the blue frame shows the interaction of the eye region and the nearby object, i.e., the ball.} \label{2}
\end{figure}

With OFDIW, we can accomplish the onfocus detection task by training the convolutional neural network (CNN) models, which is basically a classification problem. However, directly applying the conventional CNN model to classify the input image won't always work as it ignores an important factor---the eye-context interaction. As we know, eye regions provide critical cues for onfocus detection. However, by directly performing the conventional CNN model onto the input image, most of the extracted information actually comes from the context region as the eye region only occupies a very small fraction of the whole image. Although the context regions are also informative for onfocus detection, using them as the dominant cue won't obtain the correct onfocus detection results. Another strategy is to first localize the eye regions and then accomplish onfocus detection based on the features extracted on the eye regions. This strategy could obtain more accurate detection results. However, in cases when individual's gaze has the interaction with the context facial characteristics (e.g., face directions and locations) and context objects (e.g., balls or other people), the information extracted only from the eye regions is still insufficient for identifying the focus of the individual precisely (see examples in Figure~\ref{2}). Thus, designing effective frameworks to explore eye-context interaction becomes a critical issue.

To address this issue, we propose a novel network model called eye-context interaction inferring network (ECIIN), in which Context CNN extracts features with high activations on eye region, while Eye CNN extracts features with high activations on some more detailed sub-regions in eye region. Then, capsules of Context CAP and Eye CAP are learned to reflect the context status and eye status, respectively. ECIIN is characterized by: 1) It is designed with a two-stream network architecture to extract the features of the eye region and the context region separately and then infer the eye-context interaction via an interaction inference stream. 2) It has an eye region mining module, which enables it to mine the important eye regions from each input image without any specific annotation on them. 3) Considering the superior capacity of the capsule network~\cite{7} in modeling part-whole relationship and overcoming view changes, we introduce the dynamic capsule routing scheme into ECIIN to construct the eye capsules and context capsules, and meanwhile, model the interaction among the eye and context regions explicitly. To sum up, this work mainly contains three-fold contributions:

\begin{itemize}
\item {We study the novel onfocus detection task and build a large-scale dataset, i.e., OFDIW. OFDIW will be publicly available to promote further development of this research field. }
\item {We reveal the main challenges in this task and establish a novel end-to-end deep model, called ECIIN, to explore the informative eye-context interaction cues for onfocus detection.}
\item {Comprehensive experiments on the OFDIW dataset have been conducted, which benchmarks the performance of the state-of-the-art eye contact detection, gaze estimation, image categorization networks, and fine-grained image classification models, and demonstrates the superior capacity of the proposed ECIIN.}
\end{itemize}

\section{Related Works}

\textbf{Eye Contact Detection:} While this work is closely related to the field of eye contact detection~\cite{8,9,10}, the difference is that eye contact detection mainly studies the eye contact between two humans in face-to-face communication. Over there, the images or videos are usually captured by wearable imaging equipment, such as eye-tracking glasses~\cite{2,11,12}, or in the controlled image capturing setup~\cite{13}. Under this circumstance, the focus of the individual captured by the camera would be dependent on his/her interaction with the observer wearing the imaging equipment. In contrast, in the investigated onfocus detection task, we study the individual-camera eye contact. Here, the images are captured by normal cameras instead of the wearable imaging equipment. Thus, the focuses of the individuals are mainly dependent on the mental activities or physical status of themselves rather than the interaction with the observer. Therefore, the image contents presented in the study of the individual-camera eye contact as well as the messages delivered through the individual-camera eye contact would be different from those in the common human-human eye contact. To this end, we treat the detection of individual-camera eye contact as a novel task and name it as onfocus detection. Furthermore, ECD is studied in indoor environment with carefully installed equipments. The images are captured with clear background and having equal distance to the camera, thus having the constrained imaging conditions. In contrast, our task is studied in open environments with arbitrary camera locations, thus being ‘in the wild’. From this perspective, our task is beyond the exploration of ECD.

\textbf{Gaze Estimation:} Gaze Estimation~\cite{14,15,16,17} is another related topic to this work, which aims at inferring the continuous gaze directions from images or videos. Gaze estimation usually applies in determining the user's point of gaze on a display surface, such as the tablets and laptop screens, to facilitate human-machine interaction. Gaze estimation methods with this purpose can be used to provide helpful prior knowledge for eye contact detection and onfocus detection tasks and the methodologies proposed for gaze estimation can be used to identify potential eye contacts by moderate adjustments. But based on our investigation, in the studied individual-camera interaction scenario, whether the individual is onfocus or not plays a major role in the delivery of social communication signal, whereas the specific gaze direction in out of focus cases is usually without a clear subjective intention or caused by external disturbance. To this end, onfocus detection is a new research branch on Gaze Estimation. Notice that onfocus detection has its own value in practical applications. For example, when judging whether a prisoner is lying or not, one important sign is to see if the prisoner dares to look at the police in the eye. Under this circumstance, predicting eye gaze on other locations is not that important. What's more, the gaze estimation systems would spend more execution time as their algorithms are usually more complicated. Besides, as the gaze estimation methods are not designed specifically on judging onfocusness, they would have limited detection accuracy as demonstrated in the experiment section. It is worth mentioning that using the annotation of this task to assist the learning of fine gaze estimation would be another potential research direction to benefit this community. A new trend in the field of gaze estimation is to predict gaze directions for multiple persons having group activities and interactions~\cite{18,19}. Gaze360~\cite{18} uses the camera as a third perspective to observe individual interactions between groups. In contrast, we place cameras anywhere as a first perspective in the scene to study the individual-camera eye contact.

\textbf{Relevant Datasets:} There are some existing datasets established for the study of gaze estimation or eye contact detection, which are relevant to our OFDIW dataset. Among the existing datasets, Columbia Gaze~\cite{13}, Eyediap~\cite{20}, and Gaze360~\cite{18} are three widely used datasets for gaze estimation. However, these datasets are captured using controlled recording setups, not in real-world scenarios. The images in the MMDB dataset~\cite{21}, MPIIGaze dataset~\cite{22}, and GazeCapture dataset~\cite{23} are captured by eyeglasses, laptop cameras, and smartphones, respectively, in the context of face-to-face communication or human-machine(screen) interaction. Although these data are relevant to our task, they cannot be directly used to solve our problem. Take Gaze360~\cite{18} for example, it has 169,935 images but only less than 200 images (0.1$\%$) are onfocused while others are out of focus. Training on such data would be likely to bias the model towards a trivial solution for our task. This also explains why the state-of-the-art gaze models trained on the existing datasets cannot obtain good performance on our task. And that well justifies why spending efforts to establish a novel dataset for our task is necessary. Thus, these data are not suitable to be used in the study of onfocus detection. Compared to these datasets, OFDIW contains images that are captured in unconstrained real-world scenarios toward the study of individual-camera eye contact. To this end, our collected OFDIW cannot be replaced by any existing dataset to study the onfocus detection problem.

\begin{figure}[t]
  \centering
  \includegraphics[width=0.6\textwidth]{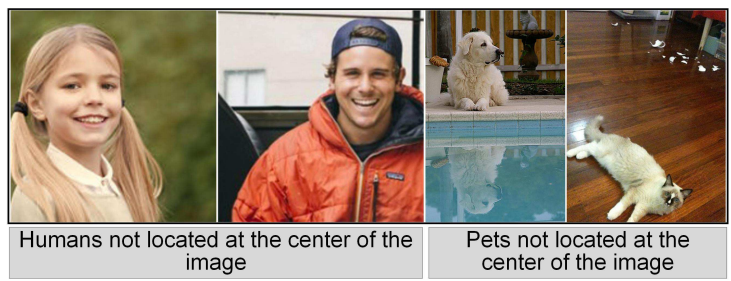}
  \caption{Illustrate the location of the subject.} \label{3}
\end{figure}

\section{The OnFocus Detection In the Wild Dataset}

\subsection{Image Collection, Split, and Ground-truth Annotation}
Our OFDIW dataset consists of two subsets: unconstrained human face subset and pet face subset, respectively. The images in the human face subset (OFDIW-HF) are acquired from the famous LFW dataset~\cite{5}, which is widely used for face recognition-related tasks. While the images in the pet face subset (OFDIW-PF) are acquired from the Oxford-IIIT Pet dataset~\cite{6}, which mainly contain faces of dog and cat. OFDIW totally contains 20,623 images, which are sufficient to implement the current deep learning-based frameworks for the onfocus detection task. Each of OFDIW-HF and OFDIW-PF is further split into the training set and the test set. The training sets of OFDIW-HF and OFDIW-PF contain 9,924 and 5,542 images, respectively. The test sets of OFDIW-HF and OFDIW-PF contain 3,309 and 1,848 images, respectively. Based on the data split, 75\% of images in OFDIW are used for training while 25\% for testing.

In order to complete the task of onfocus detection, camera can be placed anywhere in the scene to capture first-view images where the camera is on the normal line of the image plane that passes the image center. In other words, the camera is located anywhere on the subject's face, so the subject can be located anywhere in our task (see examples in Figure~\ref{3}). Notice that the investigated problem mainly deals with one subject scenario, including one-to-one or one-to-multiple communications like staff interview, president speech, prisoner interrogation, etc. Thus, in our dataset, there is only one subject appearing in the image. During the annotation, we do not use the physical location to determine whether the object is onfocus or not. Instead, we asked the annotator to use a subjective way to feel whether the subject was looking at them. For each image, there are three human annotators to label whether the individual (either the human or the pet) in the image focuses on the camera or not and obtain the final annotation by processing the majority voting on the three human annotators. The annotation processes are performed independently on different human annotators and different images.

\subsection{Diversity and Difficulty}
To reveal the diversity of the OFDIW dataset, we also conduct statistics from different aspects. Specifically, for OFDIW-HF, we conduct statistics on different age levels of the individuals presented in the images, the complexity levels of the image scenes, the occlusion of the human faces, the emotion distribution of the individuals, the gender distribution of the individuals, and the face orientation distribution of the individual. For OFDIW-PF, we conduct statistics on the distance distribution between the face of the pet to the camera, the face directions of the pets, and the breed distributions of cat and dog, respectively.

\begin{figure}[t]
	\centering
	\includegraphics[width=1\textwidth]{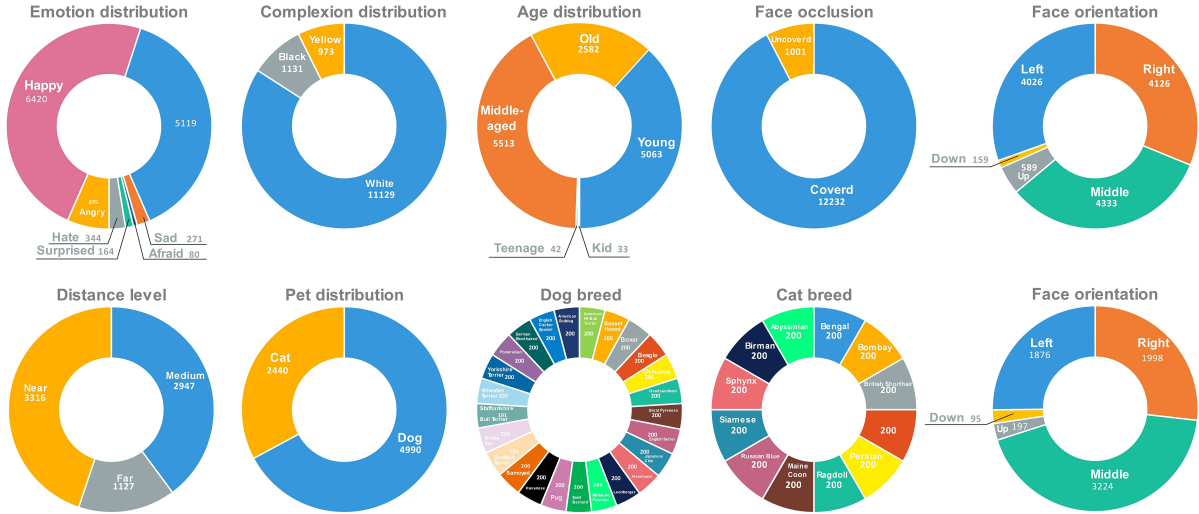}
	\caption{Diversity of the collected OFDIW dataset. Please zoom in for details.} \label{4}
\end{figure}

From the statistics results shown in Figure~\ref{4}, we can observe that OFDIW-HF exhibits a natural long-tailed distribution on emotion, age, and face orientation. In terms of emotion, most of the individuals present happy and neutral and there are also small groups of individuals present mad, sad, surprise, etc. With respect to age, most of the individuals are middle-aged and young and there are also small groups of individuals who are kids and teenagers. In terms of face orientation, most of the individuals face left, right, or middle, while a few individuals face up and down. Among these images, most of the individuals (about 85\%) are white while the rests are black and yellow. In addition, face occlusion occurs in more than 90\% images. In OFDIW-PF, about 33\% are cat images while the rests are dog images. It contains 25 dog breeds and 12 cat breeds in total, which are uniformly distributed. Among these images, most of the individuals (about 85\%) are close to or have the medium distance to the camera, while a few individuals are far from the camera. The face orientation distribution of the pets has the similar long-tailed property as that of the humans.

\section{Eye-Context Interaction Inferring Network}

Given each input image, the proposed ECIIN model detects onfocus phenomenon by automatically mining the eye regions from the image and then modeling the interaction between the eye region and the context region. Different from the previous eye contact detection works~\cite{9,24,25}, ECIIN does not require face detection, facial landmarks tracking, or face pose estimation results to make the final decision. Instead, we introduce several carefully designed network modules into ECIIN to learn to explore such information cues implicitly by casting the whole learning problem as an image categorization task.

\subsection{Network Design}

ECIIN first enables two network streams to separately extract information from the context region and the eye region, and then these two streams are fused into one stream to explore the eye-context interaction. For each of the eye region stream and the context region stream, we use two network blocks to extract useful information, i.e., the convolutional neural network (CNN) block and the capsule network (CAP) block. Based on our intuition, the CNN block is first used to extract high-dimensional region feature to replace the original RGB pixel feature. CAP block is used to explore the spatial-hierarchical information upon the regions to obtain the final prediction. In the CNN block, conventional convolution layers are adopted to extract feature maps for the input image region. Here we follow the widely-used VGG network architecture~\cite{26} to implement our CNN block by using its conv1- conv5 layers. Then, considering that Capsule network is good at exploring the part-whole relation, we  formulate the eye-context relation (a critical issue for onfocus detection) as an abstract part-whole relation in our model, where a CAP block is used to construct primary capsules based on the features extracted by the CNN block and then conduct the conv-capsule mapping to model the part-whole relationship within the input image region (either the context region or the eye region).

\begin{figure}[t]
	\centering
	\includegraphics[width=1\textwidth]{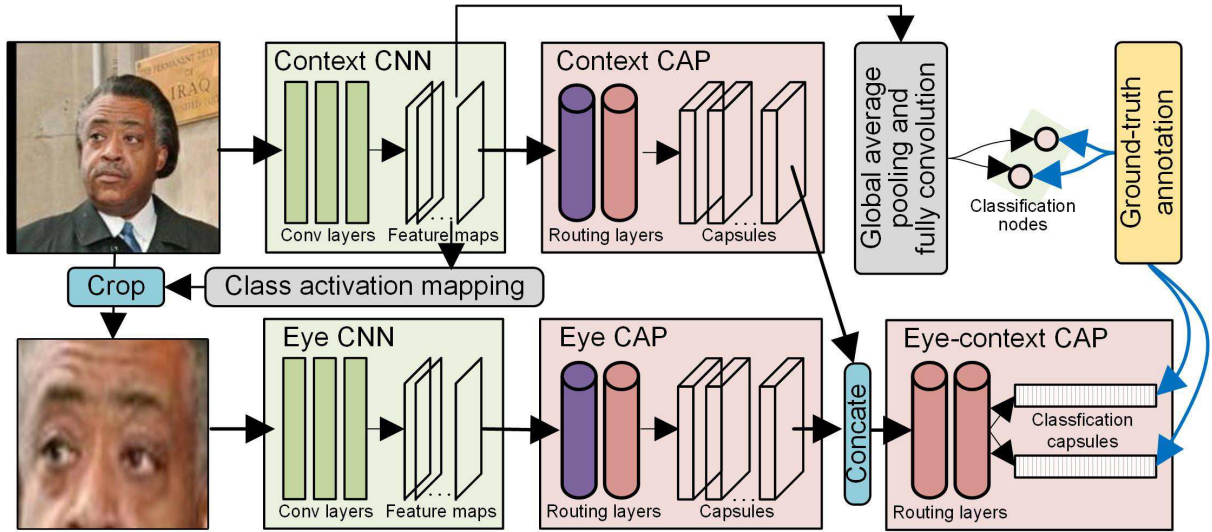}
	\caption{Framework of the eye-context interaction inferring network (ECIIN) for onfocus detection. The purple blocks indicate the primary capsule layers while the pink blocks indicate the conv-capsule layers. Blue arrows indicate the supervision information. The inputs of the two network streams are with the spatial size of $224 \times 224$. The whole network is trained in an end-to-end fashion. } \label{f}
\end{figure}

\subsection{CAP Modeling}

\begin{figure}[t]
	\centering
	\includegraphics[width=1\textwidth]{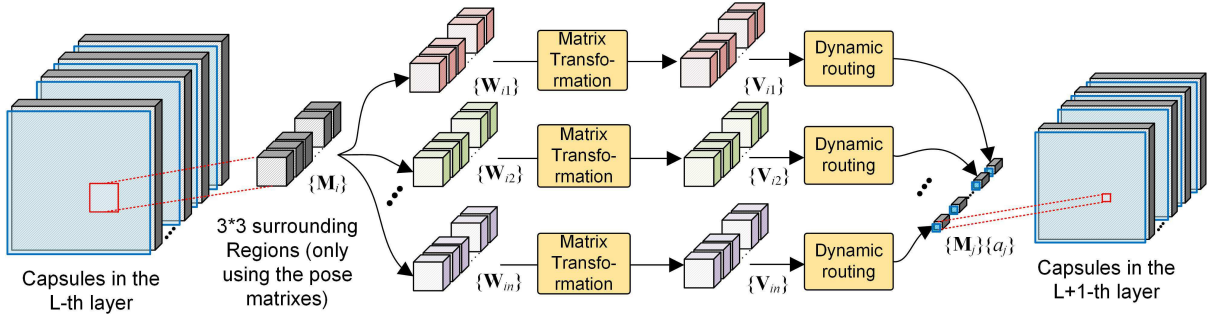}
	\caption{Illustrate the capsule convolution operation between two capsule convolutional layer, where $\{\textbf{W}_{ij}\}$ and $\{\textbf{V}_{ij}\}$ indicate the transformation matrixes and the vote matrixes, respectively. $\{\textbf{M}_{i}\}$ and $\{\textbf{M}_{j}\}$ indicate the capsules from the current layer and the next layer, respectively. The blue channel in the capsules indicates the activation while the gray channels indicate the pose matrix. } \label{r}
\end{figure}

Given feature maps $\textbf{F} \in \mathbb{R}^{(W\times H\times 512)}$ obtained from the CNN block, we construct the $4\times 4$ pose matrixes of the primary capsules by conducting $n_p$ $1\times 1$ convolutions on $\textbf{F}$ and the 1-dimensional activations of the primary capsules by applying the sigmoid function
on $\textbf{F}$. This constructs $W\times H \times \frac{n_p}{4\times 4}$ primary capsules\footnote{Here we can interpret it as $\frac{n_p}{4\times 4}$ types of capsules while each type contains $W\times H$ entities at the corresponding positions.} where each of them has a $4\times 4$ pose matrix and a 1-dimensional activation. Then, in each of the convolutional capsule layers, we set the kernel size as $k$ and the new capsule type as $n_t$, indicating that there will be $n_t$ capsules in each location of the next layer and each of these capsules will be obtained by aggregating capsules from the current layer of its $k\times k$ surrounding regions. The aggregation is performed by a matrix transformation process and a dynamic routing process. Denote the pose matrixes of the current capsule layer as $\{\textbf{M}_i\}$, while the vote matrixes of the next capsule layer as $\{\textbf{V}_{ij}\}$. The matrix transformation process is performed by learning the transformation matrixes $\{\textbf{W}_{ij}\}$ such as $\textbf{V}_{ij}=\textbf{M}_i \textbf{W}_{ij}$. The dynamic routing process is performed by the EM-based routing-by-agreement strategy~\cite{7} to obtain the capsules which consist of both the pose matrixes $\{\textbf{M}_j\}$ and the activations $\{a_j\}$ in the next layer. More specifically, for each dimension of the vectorized pose matrix $\textbf{M}_j$, we formulate it as the expectation of a Gaussian distribution on the corresponding dimension of $\{\textbf{V}_{ij}\}$. Denote the probability of $\{v_{ij}^h\}$ belonging to the Gaussian distribution of the $j$-th capsule as $p_{i|j}^h$ ($h$ is the dimension index of the vectorized pose matrix), we calculate the activation of the $j$-th capsule using the minimum description length principle~\cite{7}:
\begin{equation}
a_{j}={logistic}\left(\lambda\left(\beta_{a}-\beta_{u} \sum_{i} r_{i j}-\sum_{h} {cost}_{j}^{h}\right)\right), \ \ {cost}_{j}^{h}=-\sum_{i} r_{i j}\ln p_{i|j}^h,
\end{equation}
where $\beta_{a}$ and $\beta_{u}$ are two learnable parameters in the capsule routing process, ${cost}_{j}^{h}$ indicates the cost for using the capsules in the current layer to active the $j$-th capsule in the next layer. $r_{i j}$ is the weight for assigning the $i$-th capsule in the current layer to the $j$-th capsule in the next layer, which is initialized uniformly and then gradually updated
in the routing process. $\lambda$ is an inverse temperature parameter which increases at each iteration. An illustration of the above-mentioned convolutional capsule mapping operation is shown in Figure~\ref{r}.

As for the network layers in the eye-context CAP block, similar operations are adopted. Specifically, it takes the concatenation of the context capsules and eye capsules as the input. Then, it has two routing layers. The first one implements the conv-capsule mapping to explore the eye-context relationship among the context capsules and eye capsules. While the second one infers the two classification capsules for final prediction.

\subsection{Learning Objective Function}

\begin{figure}[t]
	\centering
	\includegraphics[width=1\textwidth]{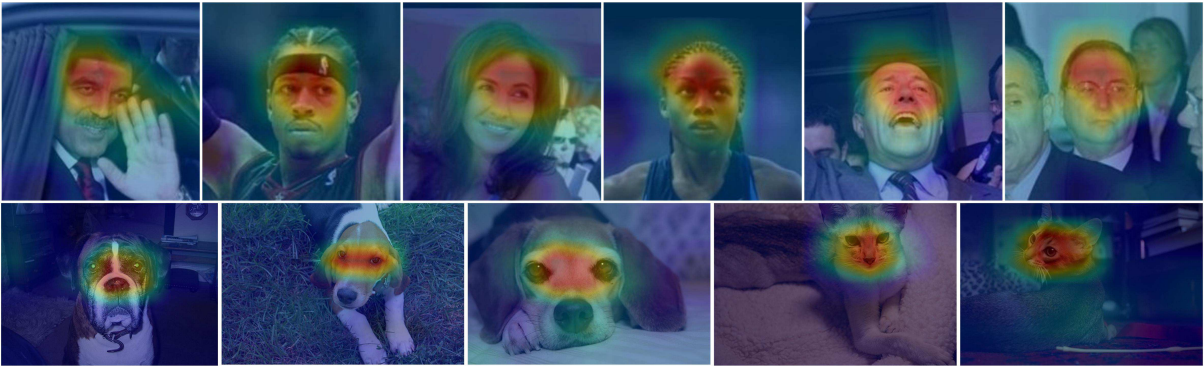}
	\caption{Some examples of the eye regions mined by our approach, where regions with red masks are predicted with higher probability for being the eye regions.} \label{eye}
\end{figure}

To train ECIIN, we follow~\cite{7} to adopt the spread loss by maximizing the gap between the activation of the target class and the activation of the other classes:
\begin{equation}
L=\sum_{p \neq t} \max (0, m-(a_{t}-a_{p}))^{2},
\end{equation}
where $a_t$ and $a_p$ indicate the activation of the target class and one another class, respectively. $m$ is the measure of the margin between a certain wrong class and the target class, which is set to 0.2 at the beginning of the training process and then linearly increases to 0.9 during the subsequent learning process. To mine the eye region from each input image automatically, we connect the context CNN block with a global average pooling layer~\cite{27} followed by a fully connected layer to predict the two-dimensional classification vector (see Figure~\ref{f}). Then, training this network branch through the cross-entropy loss would enable the network to mine discriminant image regions (through the class activation mapping technique~\cite{27}) which contribute a lot to the classification. From Figure~\ref{eye} we can observe that the mined image regions are mainly the eye regions, which conforms to the intuition that the eye region would play a very important role in the onfocus detection task.

\section{Experiments}

The experiments are implemented on both subsets of the built OFDIW dataset. On OFDIW-HF, we set $n_p=512$, $n_t=32$, kernel size $k$ as 3. The strides of the Context CAP block and Eye CAP block are set to 2 while the stride of the Eye-Context CAP block is set to 1. On OFDIW-PF, considering that there are less data for training, we reduce the model parameter to avoid overfitting. Specifically, we add a max-pooling layer after the Context CNN block and Eye CNN block, and set $n_p=256$, $n_t$ as 8 for the Context CAP block and Eye CAP block while 4 for the Eye-Context CAP block. The stride is set to 1 for all the three CAP blocks. We adopt the widely used classification accuracy and F-measure in evaluation:
\begin{equation}
\text{Accuracy}=\frac{\text{TP}+\text{TN}}{\text{TP}+\text{TN}+\text{FP}+\text{FN}}, \ \ \ \ \ \ \text{F-measure}=\frac{2 \times \text{Precision} \times \text{Recall} }{\text{Precision} + \text{Recall}},
\end{equation}
where TP, TN, FP, FN indicates the numbers of true positive, true negative, false positive, and false negative samples, respectively. Precision and Recall is calculated by $\text{Precision}=\frac{\text{TP}}{\text{TP}+\text{FP}}$, and $\text{Recall}=\frac{\text{TP}}{\text{TP}+\text{FN}}$.

\begin{table}[t]\small
\begin{minipage}{0.5\textwidth}
\captionsetup{width=0.9\textwidth}\centering
\makeatletter\def\@captype{table}
\caption{Comparison to the state-of-the-art methods, including the eye contact detection methods, gaze estimation methods, image categorization networks, and fine-grained image classification networks, on OFDIW-HF.}
\label{tab:tb1}
\begin{tabular*}{0.9\textwidth}{lcc}
    \toprule
    Methods & Accuracy & F-measure \\
    \midrule
    DEEPEC~\cite{10}  & 0.7939  & 0.8639  \\
    Gaze-lock detector~\cite{24} & 0.8129  & 0.8821  \\
    PiCNN~\cite{2} & 0.7996  & 0.8747  \\
    Multimodal CNN~\cite{14} & 0.7634  & 0.8476  \\
    CA-Net~\cite{28} & 0.8117  & 0.8809  \\
    \midrule
    NTSnet~\cite{29} & 0.8190  & 0.8834 \\
    MPN-COV~\cite{30} & 0.8362  & 0.8926  \\
    DFL~\cite{31} & 0.8102  & 0.8814  \\
    BCNN~\cite{32} & 0.8320  & 0.8919  \\
    DCL~\cite{33} & 0.8214  & 0.8867 \\
    \midrule
    VGG16~\cite{26} & 0.8262  & 0.8897  \\
    Resnet50~\cite{34} & 0.8229  & 0.8837  \\
    Res2net50~\cite{35}  & 0.8223 & 0.8861  \\
    Densenet121~\cite{36} & 0.8290  & 0.8904 \\
    Senet154~\cite{37} & 0.8193  & 0.8832 \\
    \midrule
    Ours  & \textbf{0.8471} & \textbf{0.9007} \\
    \bottomrule
\end{tabular*}
\end{minipage}
\begin{minipage}{0.5\textwidth}
\captionsetup{width=0.9\textwidth}\centering
\makeatletter\def\@captype{table}
\caption{Comparison to the state-of-the-art methods, including the eye contact detection methods, gaze estimation methods, image categorization networks, and fine-grained image classification networks, on OFDIW-PF.}
\label{tab:tb2}
\begin{tabular*}{0.9\textwidth}{lcc}
    \toprule
    Methods & Accuracy & F-measure \\
    \midrule
    DEEPEC~\cite{10} & 0.5411  & 0.6323  \\
    Gaze-lock Detector~\cite{24} & 0.5801  & 0.6450  \\
    PiCNN~\cite{2} & 0.5584  & 0.5996  \\
    Multimodal CNN~\cite{14}& 0.5487  & 0.6329  \\
    CA-Net~\cite{28} & 0.6272  & 0.6518  \\
    \midrule
    NTSnet~\cite{29} & 0.7030  & 0.7183  \\
    MPN-COV~\cite{30} & 0.7192  & 0.7435  \\
    DFL~\cite{31} &   0.7495  & 0.7504  \\
    BCNN~\cite{32} & 0.7608  & 0.7600  \\
    DCL~\cite{33} & 0.7576  & 0.7679 \\
    \midrule
    VGG16~\cite{26} & 0.7516  & 0.7536  \\
    Resnet50~\cite{34} & 0.7446  & 0.7495 \\
    Res2net50~\cite{35}  & 0.7505  & 0.7572  \\
    Densenet121~\cite{36} & 0.7608  & 0.7676  \\
    Senet154~\cite{37} & 0.7673  & 0.7746  \\
    \midrule
    Ours  & \textbf{0.7744} & \textbf{0.7747} \\
    \bottomrule
\end{tabular*}
\end{minipage}
\end{table}

\begin{table}\small
\begin{minipage}{0.6\textwidth}
\captionsetup{width=0.9\textwidth}\centering
\makeatletter\def\@captype{table}
\caption{Ablation study of our ECIIN model on OFDIW-HF. ``2S'' is short for two-stream, ``EC CAP'' is short for eye-context capsule block.}
\label{tab:tb3}
\begin{tabular*}{0.9\textwidth}{lcc}
    \toprule
    Methods & Accuracy & F-measure \\
    \midrule
    2S CNN & 0.8244 &	0.8890 \\
    2S CNN + Context CAP & 0.8302 &	0.8881 \\
    2S CNN + 2S CAP & 0.8401 &	0.8964  \\
    2S CNN + 2S CAP + EC CAP & 0.8471 &	0.9007 \\
    \bottomrule
\end{tabular*}
\end{minipage}
\begin{minipage}{0.4\textwidth}
\captionsetup{width=0.9\textwidth}\centering
\makeatletter\def\@captype{table}
\caption{Study on different strategies to generate the input eye regions of the Eye CNN block.}
\label{tab:tb4}
\begin{tabular*}{0.9\textwidth}{lcc}
    \toprule
    Strategies & Accuracy & F-measure \\
    \midrule
    WM  & 0.8395  & 0.8961 \\
    CO & 0.8419  & 0.8970  \\
    BM & 0.8465 & 0.8990  \\
    CR &0.8471  & 0.9007  \\
    \bottomrule
\end{tabular*}
\end{minipage}
\end{table}

\subsection{Comparison to State-of-the-Art Methods }

In this section, we first compare our approach with several existing eye contact detection and gaze estimation methods~\cite{2,10,14,24,28}. When compared with the gaze estimation methods, we simply modify the prediction head of their models for two-class classification. It is also worth mentioning that most eye contact detection and gaze estimation methods use the pre-trained facial landmark detector to locate the eye and other facial components from each input image firstly and then train the learning models. In contrast, our proposed approach directly performs on the raw input image without requiring the pre-trained facial landmark detector. The experimental results on two subsets of OFDIW are reported on the top blocks of Table~\ref{tab:tb1} and Table~\ref{tab:tb2}. From the comparison results on OFDIW-HF, it can be observed that our method obtains 4.21-10.96\% relative performance gains under the measurement of detection accuracy, and 2.11-6.26\% relative performance gains under the measurement of F-measure. While such improvements become stronger on OFDIW-PF. Based on our understanding, the reason for the failure of the eye contact detection and gaze estimation methods on OFDIW-PF might be that these methods cannot obtain precise locations of the eye and other facial components of pets. This demonstrates that our approach can work flexibly to detect individual-camera eye contact for both human and pets.

\begin{figure}[t]
	\centering
	\includegraphics[width=1\textwidth]{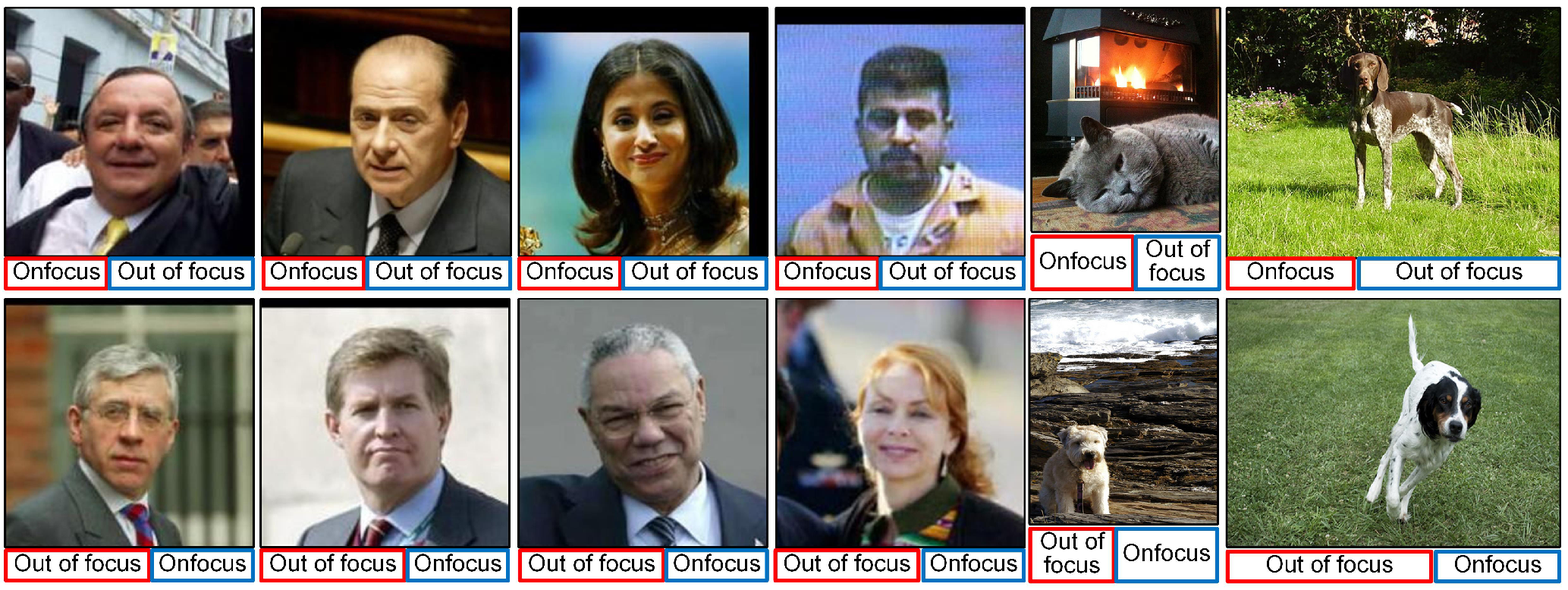}
	\caption{Some examples to illustrate the failure cases of the proposed approach, where the ground-truth labels are shown in the red boxes while the prediction results are shown in the blue boxes.}
\label{failure}
\end{figure}

We also compared our approach with the state-of-the-art image categorization methods~\cite{26,34,35,36,37} and fine-grained image classification methods~\cite{29,30,31,32,33}. The corresponding comparison results are reported in the middle and bottom blocks of Table~\ref{tab:tb1} and Table~\ref{tab:tb2}, respectively. Notice that both our model and the compared models use the parameters pre-trained on the ImageNet dataset~\cite{38}. From the comparison results, we can observe that directly using these state-of-the-art network architectures cannot obtain better performance than our proposed ECIIN model. Take the results on OFDIW-HF as an example. When comparing to the state-of-the-art image categorization networks and fine-grained image classification networks, our approach obtains 1.26-4.55\% relative performance gains under the detection accuracy. It worth mentioning that during the comparison of classification models, such improvements are not marginal as according to the current literature, e.g., \cite{40,41}, a new classification model usually exceeds the existing models around 1\% in accuracy. To our best knowledge, the advantage of our approach upon the existing image categorization models is that the proposed framework contains a two-stream architecture to model the eye-context interaction. While compared to the fine-grained classification models, the advantage of the proposed approach is the capacity to facilitate the part-whole relationship modeling for onfocus detection.

For model complexity, when compared to the gaze models, the complexity of our model is lower as gaze models usually need to integrate multiple deep models to accomplish the task. When compared to the compared classification networks, our model has higher complexity due to the two-stream architecture and the capsule routing layers.

To analyze the failure cases of the proposed approach, we show examples in Figure~\ref{failure}. As can be seen, the failure cases mainly occur when one or two eyeballs of the subject (either human or pet) are hard to observe from the input image. Under this circumstance, the extracted feature cannot provide sufficient information to make precise predictions.

\subsection{Ablation Study}

In the ablation study, we first study the key components of the proposed ECIIN model. Specifically, we compare the proposed approach with three baseline models. The first one is the ``2S CNN'' model, which uses the two-stream CNN architecture with the decision-level fusion. In other words, the final decision of this baseline is obtained by averaging the predictions of the two CNN streams rather than applying a fusion block to integrate the features learned from the two CNN streams to generate the final prediction. The ``2S CNN + Context CAP'' baseline adds the Context CAP on top of ``2S CNN'', while the ``2S CNN + 2S CAP'' baseline adds the Eye CAP on top of ``2S CNN + Context CAP''. Similar to ``2S CNN'', these two baselines also adopt the decision-level fusion strategy to obtain the final prediction. Finally, our approach is implemented by further adding the Eye-Context CAP on top of ``2S CNN + 2S CAP''. From the experimental results reported in Table~\ref{tab:tb3}, we can observe that using the capsule blocks can not only help better explore the information within the context image regions or the eye regions but also  improve the final prediction by exploring the interaction between the eye and context regions.

In addition, we also study several strategies to generate the input eye regions of the Eye CNN block. The studied strategies include the weighted multiplying strategy (WM), the concatenation strategy (CO), the binary multiplying strategy (BM), and the cropping strategy (CR). Specifically, the WM strategy directly implements the element-wise product between the original image and the class activation map (CAM) obtained from the Context CNN to generate Eye CNN's input. The CO strategy concatenates the original image and the class activation map as the input of the Eye CNN. The BM strategy first binarizes the obtained CAM with the threshold of 0.8 and then implements the element-wise product between the original image and the binary CAM to generate Eye CNN's input. The CR strategy uses the same threshold to binarize the CAM and then crops the rectangle image region from the original image according to the highlight locations on CAM. The experimental results reported in Table~\ref{tab:tb4} demonstrate that the CR strategy works better than other strategies in terms of both Accuracy and F-measure. To our knowledge, this might due to the CR strategy can make the Eye CNN and Eye CAP better focus on the fine-level features around the eye regions.

\section{Conclusion}
This work studies a new task called onfocus detection, which aims at identifying whether the focus of the individual (human or animal) captured by a camera in unconstrained imaging conditions is on the camera or not. To solve this problem, we first build a large-scale onfocus detection dataset, named OFDIW, to facilitate the study in this field. The OFDIW dataset contains individuals with diverse emotions, ages, facial characteristics, and rich interactions with surrounding objects and background scenes. It consists of two subsets and totally 20,623 unconstrained images. Then, we reveal the main challenges in the onfocus detection task and establish a novel end-to-end deep model, called ECIIN, to explore the informative eye-context interaction cues for onfocus detection. Comprehensive experiments on the OFDIW dataset demonstrate that the proposed model can obtain superior performance when compared to the state-of-the-art gaze estimation, eye contact detection, image categorization and fine-grained classification methods. Besides, comprehensive ablation studies are also conducted to verify the effectiveness of each component considered in our approach.

\Acknowledgements{This work was supported in part by the National Science Foundation of China (Grant Nos. 61876140 and 61773301), the Fundamental Research Funds for the Central Universities (Grant Nos. JBZ170401) and the China Postdoctoral Support Scheme for Innovative Talents (Grant Nos. BX20180236).}




\end{document}